\documentclass[conference]{IEEEtran}
\IEEEoverridecommandlockouts
\usepackage{cite}
\usepackage{lipsum} 
\usepackage{amsmath,amssymb,amsfonts}
\usepackage{algorithmic}
\usepackage{float}
\usepackage{graphicx}
\usepackage{textcomp}
\usepackage{xcolor}
\usepackage[margin=2cm]{geometry}
\usepackage{xcolor}
\usepackage[most]{tcolorbox}
\IEEEtriggeratref{18}
\usepackage[switch]{lineno} 
\usepackage[utf8]{inputenc}

\begin{document}

\title{Exploring the Efficacy of Robotic Assistants with ChatGPT and Claude in Enhancing ADHD Therapy: Innovating Treatment Paradigms}

\makeatletter
\newcommand{\linebreakand}{%
  \end{@IEEEauthorhalign}
  \hfill\mbox{}\par
  \mbox{}\hfill\begin{@IEEEauthorhalign}
}
\makeatother

\author{\IEEEauthorblockN{Santiago Berrezueta-Guzman}
\IEEEauthorblockA{
	\textit{Technical University of Munich}\\
	s.berrezueta@tum.de}
\and
\IEEEauthorblockN{Mohanad Kandil}
\IEEEauthorblockA{
	\textit{Technical University of Munich}\\
	mohanad.kandil@tum.de}
\and
\IEEEauthorblockN{María-Luisa Martín-Ruiz}
\IEEEauthorblockA{
	\textit{Universidad Politécnica de Madrid}\\
	marialuisa.martinr@upm.es}
\linebreakand
\IEEEauthorblockN{Iván Pau de la Cruz}
\IEEEauthorblockA{
	\textit{Universidad Politécnica de Madrid} \\
	ivan.pau@upm.es}
\and
\IEEEauthorblockN{Stephan Krusche}
\IEEEauthorblockA{
	\textit{Technical University of Munich}\\
	krusche@tum.de}
}

\maketitle

\begin{abstract}

Attention Deficit Hyperactivity Disorder (ADHD) is a neurodevelopmental condition characterized by inattention, hyperactivity, and impulsivity, which can significantly impact an individual's daily functioning and quality of life. Occupational therapy plays a crucial role in managing ADHD by fostering the development of skills needed for daily living and enhancing an individual's ability to participate fully in school, home, and social situations. 
Recent studies highlight the potential of integrating Large Language Models (LLMs) like ChatGPT and Socially Assistive Robots (SAR) to improve psychological treatments. This integration aims to overcome existing limitations in mental health therapy by providing tailored support and adapting to the unique needs of this sensitive group. However, there remains a significant gap in research exploring the combined use of these advanced technologies in ADHD therapy, suggesting an opportunity for novel therapeutic approaches.

Thus, we integrated two advanced language models, ChatGPT-4 Turbo and Claude-3 Opus, into a robotic assistant to explore how well each model performs in robot-assisted interactions. Additionally, we have compared their performance in a simulated therapy scenario to gauge their effectiveness against a clinically validated customized model. 
The results of this study show that ChatGPT-4 Turbo excelled in performance and responsiveness, making it suitable for time-sensitive applications. Claude-3 Opus, on the other hand, showed strengths in understanding, coherence, and ethical considerations, prioritizing safe and engaging interactions. Both models demonstrated innovation and adaptability, but ChatGPT-4 Turbo offered greater ease of integration and broader language support. The selection between them hinges on the specific demands of ADHD therapy.

\end{abstract}

\begin{IEEEkeywords}
Artificial Intelligence, LLMs, Cognitive Therapy, ADHD, ChatGPT, Customizable AI Bots, Robotic Systems in Therapy, Occupational Therapy Innovation, Personalized Therapy Sessions, AI in Mental Health.
\end{IEEEkeywords}

\section{Introduction}

Attention-Deficit/Hyperactivity Disorder (ADHD) is a neurodevelopmental disorder characterized by patterns of inattention, hyperactivity, and impulsivity that are pervasive, impairing, and inconsistent with the developmental level of a person \cite{fletcher2021attention}. The Diagnostic and Statistical Manual of Mental Disorders (DSM-V) states that for a diagnosis, symptoms should appear before age 12, be noticeable in multiple settings like home and school, and significantly affect daily functioning \cite{koutsoklenis2023adhd}, \cite{bell2011critical}. 

Non-pharmacological treatments for ADHD are increasingly recognized for their value as complementary or alternative options to conventional medication. These interventions, which include behavioral therapies, cognitive training, dietary modifications, and exercise, are designed to mitigate ADHD symptoms and enhance overall functioning \cite{lambez2020non, healthcare11202795}. Among various approaches, occupational therapy is notable for focusing on methods to manage behavior and encourage positive actions \cite{westwood2023computerized}.

Additionally, it is crucial to consider technology's evolving role in mental health care. Integrating Artificial Intelligence (AI) in the field presents a new avenue worth examining as we navigate various treatment options. Embracing AI in mental health therapy introduces a spectrum of opinions and considerations among patients and professionals. The question of whether people prefer AI to traditional mental health therapy reveals a complex landscape, with varied perspectives emerging from recent research. Some studies suggest apprehension towards AI in mental health due to concerns like data privacy, empathy, and successful user integration, indicating a preference for human psychologists in specific contexts \cite{oladimeji2023impact}. Conversely, individuals experiencing concerns related to stigma may prefer AI-powered virtual therapists due to the perceived secure environment, continuous support, and reduced barriers to seeking help \cite{darzi2023could}. This is particularly relevant for mental health therapy, where AI's ability to tailor solutions, monitor progress efficiently, and analyze data from various sources shows promise in augmenting traditional therapeutic practices.

Therefore, this research investigates a novel approach to therapy for children with ADHD, focusing on enhancing personalization and interactivity. We evaluated the integration of \textit{ChatGPT-4 Turbo}\footnote{ChatGPT-4 Turbo is a highly optimized version of the ChatGPT-4 model, designed for faster response times and improved efficiency in generating text \cite{achiam2023gpt}.} and \textit{Claude-3 Opus}\footnote{Claude-3  is an AI model designed for conversational understanding, focusing on generating coherent and contextually relevant responses. It aims to enhance user interactions through improved ethical alignment and reduced biases \cite{cowen2024claude}.} within a Robotic Assistant. This integration underwent technical and clinical evaluation, and the findings indicate promising potential for offering personalized and interactive therapies. These advancements are expected to significantly enhance engagement and therapeutic outcomes for children with ADHD.

This paper comprises Section \ref{RW}, which reviews related work in the field of  LLMs and robotic assistants supporting ADHD. The subsection \ref{bk} explains our previous work using robotic assistants for ADHD therapies and the limitations these LLMs could overcome, such as enhancing the interactivity of therapies. Section \ref{M} explains how ChatGPT-4 Turbo and Claude-3 were integrated and tested with the robotic assistant. The results of this research are presented in Section \ref{R} and discussed in Section \ref{D}. The conclusions and the future challenges for research in this area are in section \ref{C}.

\section{Related work}\label{RW}

The potential applications and current limitations of ChatGPT and other LLMs opened a considerable discussion in psychiatry. In their research, Szu-Wei Cheng et al. explored how ChatGPT and other GPT models could change the field of mental health care. They pointed out that while these technologies could greatly help with everyday tasks in psychiatry, like writing medical records, improving communication, and helping with research, there are still some challenges to overcome before they can be fully used in practice. They emphasized the need for clear ethical guidelines to ensure these tools are used safely and effectively in all areas of mental health care, including online therapy and educational settings. However, the team imagines a future where these technologies could independently conduct a complete therapy session, understanding and responding to human emotions and needs \cite{cheng2023now}. 

In the same way, the potential applications and current limitations of socially assistive robotics (SAR) in psychiatry have been discussed deeply in the last few years. Rabbitt et al. researched how many people need mental health services but don't get them. They believe robots can benefit mental health care, discussing their current use and potential future applications. The study pointed out that these robots can help reach people who don't usually get enough support and can provide help consistently and repeatedly. But, some issues need to be worked out, like dealing with privacy and ethical questions and overcoming some tech hurdles. They think these robots can meet unmet needs in mental health care, but they also remind us that robots should not replace real professionals in providing care \cite{rabbitt2015integrating}.

In the specific field of ADHD, a notable study by Tamdjidi et al. investigated ChatGPT as an assistive technology in reading comprehension for individuals with ADHD \cite{Tamdjidi1778288}. Participants with and without ADHD were assessed through reading comprehension tests conducted with and without the assistance of ChatGPT. 
The intriguing findings revealed an interesting pattern among participants with prior ChatGPT experience who increased their comprehension abilities. This suggests that familiarity with the tool significantly affects its effectiveness as an assistive technology.

On the other hand, robotics has significantly influenced the support and treatment of individuals with ADHD, offering novel and effective methods for addressing various aspects of the condition. The utilization of robotics has been instrumental in facilitating language development \cite{estevez2021case}, providing therapeutic interventions \cite{amato2021socially, lai2021data, arpaia2020wearable, rakhymbayeva2020long, zhanatkyzy2020quantitative}, supporting learning processes \cite{kumazaki2021enhancing, berrezueta2021assessment}, and improving attention and memory \cite{vita2019neurobot}. These developments show how robotics can offer customized and engaging therapy options for people with ADHD. This enhances therapy approaches and opens new possibilities for research and practical use \cite{berrezueta2021robotic}.

These previous studies underscore the significance of combining LLMs with SAR to enhance and address the shortcomings of existing psychological treatments. Specifically, within the context of ADHD therapy, there is a notable absence of studies that integrate these technologies to support this vulnerable group, offering a new kind of treatment that adapts to their environment and supports them.   

\subsection{Preliminary Developed Work}\label{bk}

\begin{figure*}[h!]
	\centering
	\includegraphics[width=1.0\linewidth]{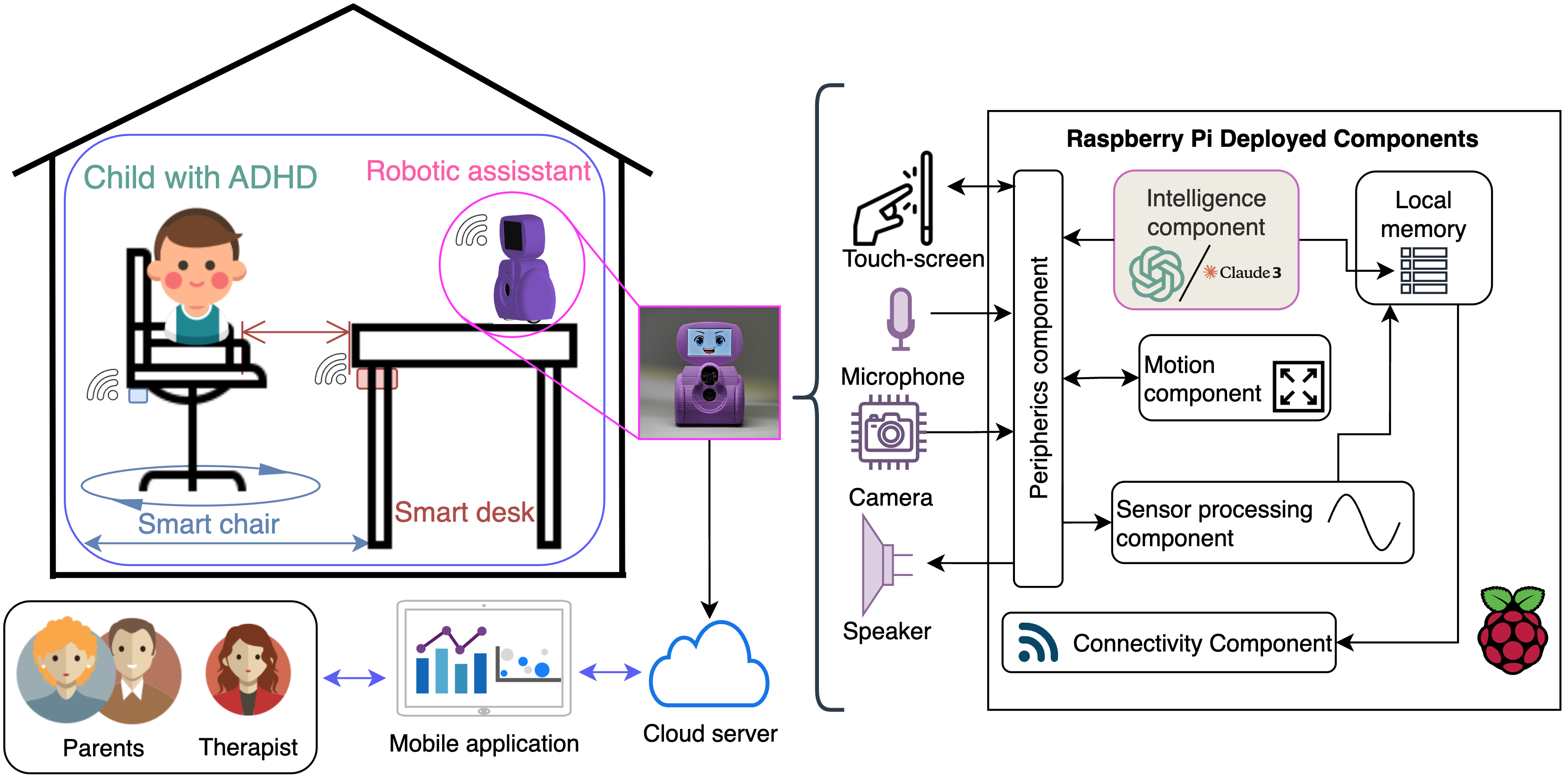}
	\caption{Illustration of the smart environment designed to facilitate occupational therapy for children with ADHD.  The detailed view on the right demonstrates the robotic assistant's incorporation of the ChatGPT model, a key innovation presented in this study.}
	\label{fig:previous}
\end{figure*}

In our earlier research, we developed a smart-home environment that helps children with ADHD maintain focus on their homework. This intelligent environment employs sensors around the child's desk and chair to detect signs of distraction, such as playing with the chair or moving away from the desk area  \cite{berrezueta2020smart, berrezueta2022design, dolon2020creation, lopez2020development }. Additionally, it includes a robotic assistant equipped with a camera that uses image recognition to identify more subtle distractions, including playing with items on the desk or daydreaming \cite{berrezueta2022artificial, berrezueta2022user}. 

However, our study found that the robot eventually became predictable and less attractive. This was due to the robot repeatedly using the same recorded messages and instructions, which made it less effective at keeping the children engaged \cite{berrezueta2021assessment}. 

Therefore, we explored integrating an LLM into our robotic assistant's interaction system to address this issue. We explored integrating a custom ChatGPT model into simulated ADHD therapies, and by using the \textit{Delphi method}\footnote{The Delphi method is a forecasting and decision-making process that gathers consensus from a panel of experts through multiple rounds of questionnaires, with summaries provided after each round to refine opinions \cite{humphrey2020delphi}.}, experts assessed the custom ChatGPT's empathy, adaptability, engagement, and communication strengths. The findings indicated that this customized ChatGPT could greatly enhance ADHD therapy with personalized therapies \cite{healthcare12060683}. 

Incorporating an LLM into our robotic assistant not only pretends to enhance its interactive capabilities but also promises to refine the therapeutic process, offering a more engaging and practical experience for children undergoing therapy. 

Figure \ref{fig:previous} presents on the left a comprehensive overview of our smart home research for supporting ADHD therapies at home, while on the right, it presents the proposed enhancement to the robotic assistant by integrating LLM into its intelligence framework. This update aims to significantly boost the assistant's ability to understand and respond to user commands and queries more effectively, leveraging the robotic's cutting-edge natural language processing capabilities. By this incorporation, the robotic assistant will become more adept at interpreting user needs, facilitating a smoother, more intuitive interaction within the smart home environment and with the child. 

\section{Methodology}\label{M}

Ensuring our robot assistant responds reliably, much like a human, is critical to avoiding the delays often experienced with traditional voice bots. Therefore, we identified two LLMs that are particularly effective for understanding complex inputs with minimal prior training, creating more realistic voice-to-voice interactions.

\subsection{Integration of the LLMs in the Robot}

We opted to incorporate GPT-4 Turbo and Claude-3 Opus into our robot. We recognized them as the premier LLMs currently available to address our testing requirements with third-party services, such as voice generation services. The comprehensive developer documentation and Application Programming Interfaces (APIs) provided by OpenAI and Anthropic significantly streamlined the process of constructing and deploying our prototype, facilitating development efficiently.

We utilized the Software Development Kits (SDKs) provided by OpenAI (for ChatGPT-4 Turbo) and Anthropic (for Claude-3). These SDKs facilitated a reliable deployment and allowed multiple testing phases within our research development and validation. Additionally, we designed a simplistic animated face that simulates human-like speech movements in sync with the audio stream being played. This integration enhanced the robot's interactivity and enriched the user experience by providing a more engaging and \textit{"humanized"} interface.

This implementation was developed in a Python script. It was tested on a Raspberry Pi, the heart of our robotic assistant, demonstrating its feasibility and seamless integration into our robotic framework.

\subsection{Integration of the speak-speak feature in the Robot}

The implementation of the speak-speak feature involves specifying the models, prompts (as we are testing in a zero learning shot environment), and a reliable architecture as illustrated in Figure \ref{fig:calude_gpt} with third-party services for voice generation, such as ElevenLabs\footnote{ElevenLabs supports the creation of realistic and expressive digital voices from text inputs. The technology behind this emphasizes the lifelike reproduction of voice and the ability to capture emotional tones and nuances, making digital communication more personal and expressive \cite{ElevenLabs2023}.}.

We chose ElevenLabs for AI voice generation because its ability to create voices in multiple languages and overcome language barriers is precious.

By efficiently incorporating LLMs, our robot can generate text—converting approximately 250 tokens into 1,000 characters. This capability enables us to effectively use ElevenLabs' starter plan, generating AI voices as user feedback of up to 30,000 characters. A critical component of our development process was using ElevenLabs' Python SDK, which empowered us to create a working prototype. Furthermore, the SDK streamlined the integration of ElevenLabs' advanced features, such as real-time streaming. 

The feature's core is to get input from a user by voice; then, the voice is converted into text and forwarded to one of the LLMs services by APIs (ChatGPT or Claude). Afterward, the output of any LLM service will stream into the ElevenLabs voice engine, which can start the voice simultaneously as it receives the first data chunk from LLM's API. The program can take 3 or 4 seconds without using streams to generate an MP3 file to play the generated voice. However, we can create an output of 600ms (milliseconds) on average by using streams, which is nearly like real-time. 
Moreover, an on-premise cloud solution can drop this latency to 150-200ms with high throughput, which is one of our future work.

\begin{figure}[h!]
	\centering
	\includegraphics[width=1.0\linewidth]{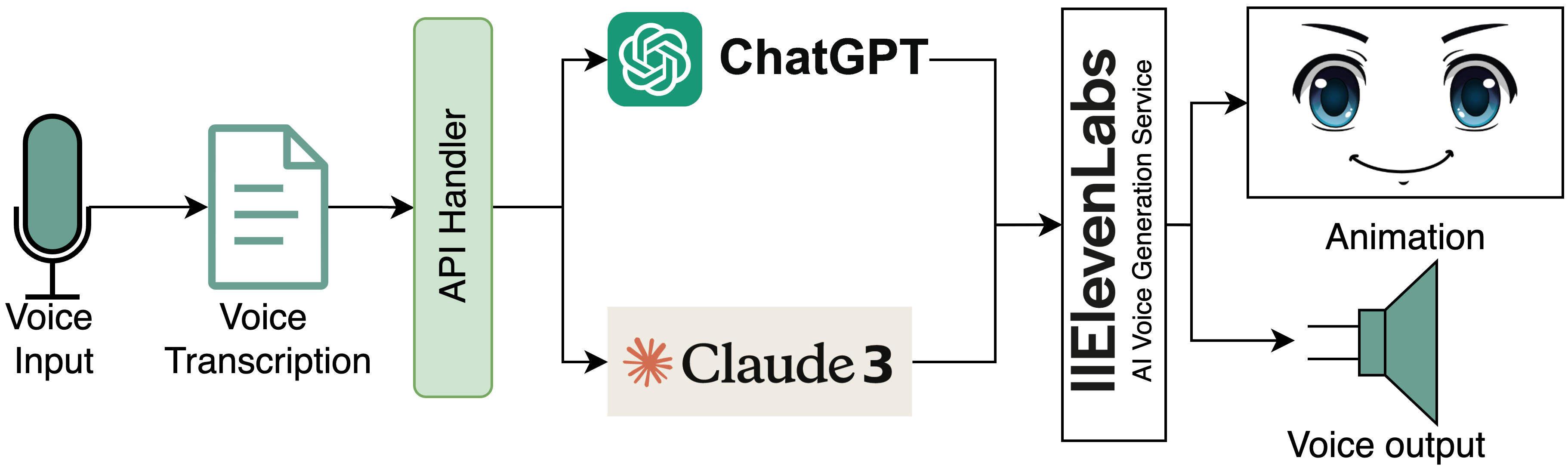}
	\caption{The system architecture where the microphone gets voice (input), then the voice gets transcribed and passed to LLMs, and the output voice is streamed to the user by the speaker. Everything is deployed inside the Raspberry Pi-4.}
	\label{fig:calude_gpt}
\end{figure}

This implementation underwent comprehensive testing to assess its viability in terms of both technical performance and clinical applicability. Figure \ref{fig:robot} displays the custom-built robotic assistant for implementing the LLMs. This robot was created to satisfy the specific hardware needs of our study, enabling effective execution of the test. Its design ensures compatibility with the computational demands of LLM operations and supports ongoing research adaptability. The results are detailed in the section below.

\begin{figure}[h!]
	\centering
	\includegraphics[width=1.0\linewidth]{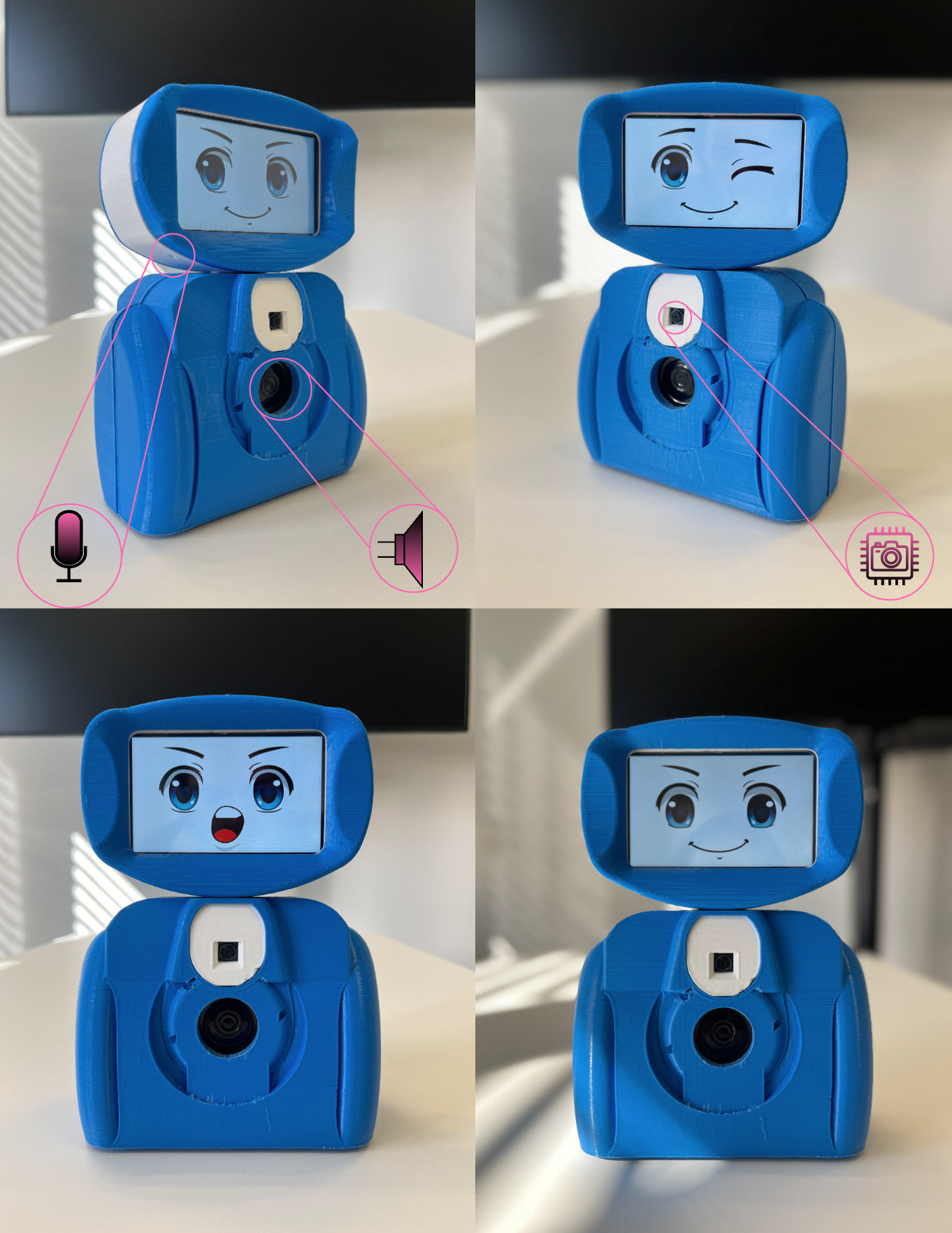}
	\caption{The used robotic assistant integrated with the selected LLMs. The pictures showcase a variety of expressions and the robot design. The peripheral components, such as the microphone, speaker, camera, and screen showing facial expressions when it talks, are also highlighted.}
	\label{fig:robot}
\end{figure}

\subsection{Technical Validation Method}\label{TV}

We conducted an in-depth technical evaluation of ChatGPT-4 Turbo and Claude-3 Opus using qualitative and quantitative methods. This involved testing a variety of inputs to check how accurate the models were, timing their responses to determine their efficiency, and engaging them in long conversations to evaluate how well they could keep up a talk coherently. We also looked at how flexible they were by changing up the prompts and finally tested how easily they could be incorporated with the existing features of our robotic system. 

\subsection{Clinical Validation Method}\label{CV}

In the clinical part of our study, we used the \textit{Delphi method} for a qualitative evaluation of our robotic assistant during mock therapy sessions, with insights from professionals. A group of ten therapists reviewed the assistant's performance, powered by ChatGPT-4 Turbo and Claude-3 Opus, and functioned in English and Spanish. They examined the assistant on several counts, such as emotional responsiveness, communication skills, ability to engage, adaptability during therapy sessions, and other sophisticated therapeutic criteria. These findings were then compared with those from a customized ChatGPT model that had previously undergone the same evaluation process by the therapists but using a computer only \cite{healthcare12060683}.  

\section{Results}\label{R}

In our study, we conducted technical and clinical tests to evaluate how well ChatGPT-4 Turbo and Claude-3 work in a therapy setting for ADHD. The evaluation by the experts was graded on a scale from 0 to 5, where 0 is the worst grade, and 5 is the maximum. Additionally, all the figures in this section were adjusted on the Y-axis to present only relevant data.

\subsection{Technical Validation Results}\label{TVR}
The technical tests looked at how the models perform, how quickly they respond, and how easily they can be integrated. Such a holistic assessment helped us identify each model's strengths and limitations illustrated in Figure \ref{fig:technicalvalidation}. 

\begin{figure}[h!]
	\centering
	\includegraphics[width=1.0\linewidth]{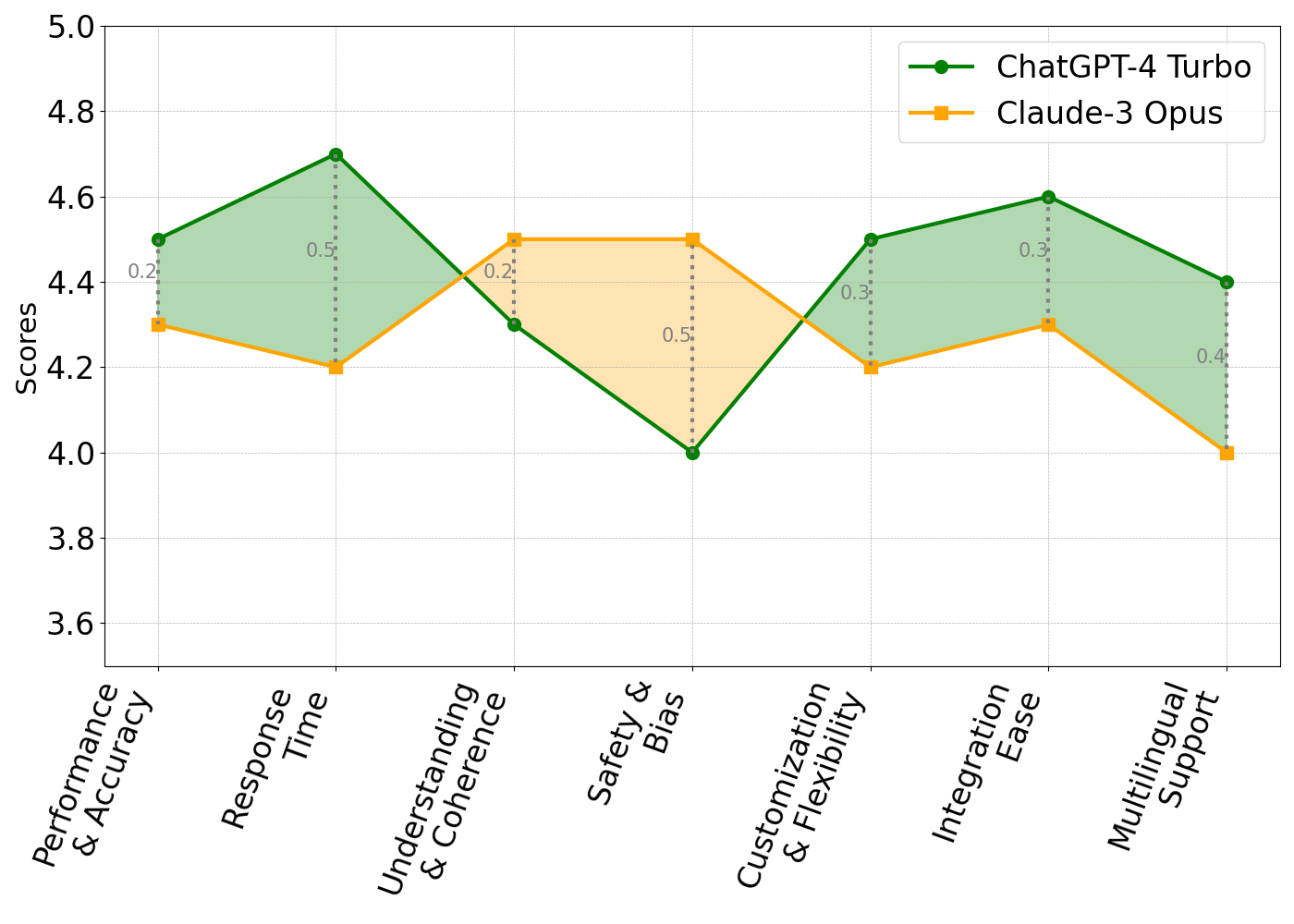}
	\caption{Comparative analysis of ChatGPT-4 Turbo and Claude-3 Opus across seven key technical parameters. Differences between the models are visualized, underscoring distinct strengths and improvement opportunities, thus providing a nuanced overview of their comparative capabilities.}
	\label{fig:technicalvalidation}
\end{figure}

\begin{itemize}
\item \textit{Performance and Accuracy:} ChatGPT-4 Turbo slightly outperforms Claude-3 Opus with a score of 4.5, offering highly accurate responses due to its broad dataset training. Claude-3 Opus, with a score of 4.3, also delivers strong performance, with a slight focus on safety, potentially limiting its output on unrestricted queries.
	
	\item \textit{Response Time:} ChatGPT-4 Turbo excels with a score of 4.7, optimized for rapid responses essential for real-time interactions. Claude-3 Opus, scoring 4.2, shows a deliberate processing approach influenced by safety and bias considerations, impacting speed.
	
	\item \textit{Understanding and Coherence:} Claude-3 Opus, scoring 4.5, surpasses ChatGPT-4 Turbo in maintaining conversational flow and context, critical for engaging dialogues. ChatGPT-4 Turbo, with a score of 4.3, also showcases strong coherence and understanding.
	
	\item \textit{Safety and Bias:} Claude-3 Opus leads with a score of 4.5, highlighting its commitment to ethical AI by prioritizing safety and reducing biases. ChatGPT-4 Turbo, with a 4.0 score, indicates a need for improvement.
	
	\item \textit{Customization and Flexibility:} ChatGPT-4 Turbo stands out with a score of 4.5, demonstrating significant adaptability and customization capabilities. Claude-3 Opus, scoring 4.2, remains versatile yet slightly constrained by its ethical and safety focus.
	
	\item \textit{Integration Ease:} ChatGPT-4 Turbo, scoring 4.6, shows superior ease of integration into existing systems, backed by comprehensive support. Claude-3 Opus, with a 4.3 score, also supports effective integration but may offer less flexibility.
	
	\item \textit{Innovation:} Both models score a 4.5, reflecting their ongoing dedication to advancing AI technology through continuous research and development.
	
	\item \textit{Multilingual Support:} ChatGPT-4 Turbo, with a 4.4 score, provides extensive language support, enhancing global usability. Claude-3 Opus, scoring 4.0, is competent but may have more limited language capabilities.
	
\end{itemize}

\subsection{Clinical Validation Results}\label{CVR}

The clinical tests checked how effective these models are in simulated therapy sessions, according to therapists. The findings are detailed below. 

The evaluation results of the emotional understanding and empathy aspects are shown in Figure \ref{fig:01emotional}. The customized ChatGPT generally shows consistently high performance with slight variations, indicating robustness across different emotional metrics. Claude-3 Opus excels in "Validation of Patient’s Experiences and Emotions," presenting a particular strength in recognizing and affirming emotional states, possibly due to its design focused on empathy. ChatGPT-4 Turbo, while performing well, shows a lower score in "Consistency and Appropriateness of Empathy" than Claude-3 Opus, hinting at areas for enhancement in delivering contextually appropriate empathetic responses. 

\begin{figure}[h!]
	\centering
	\includegraphics[width=1.0\linewidth]{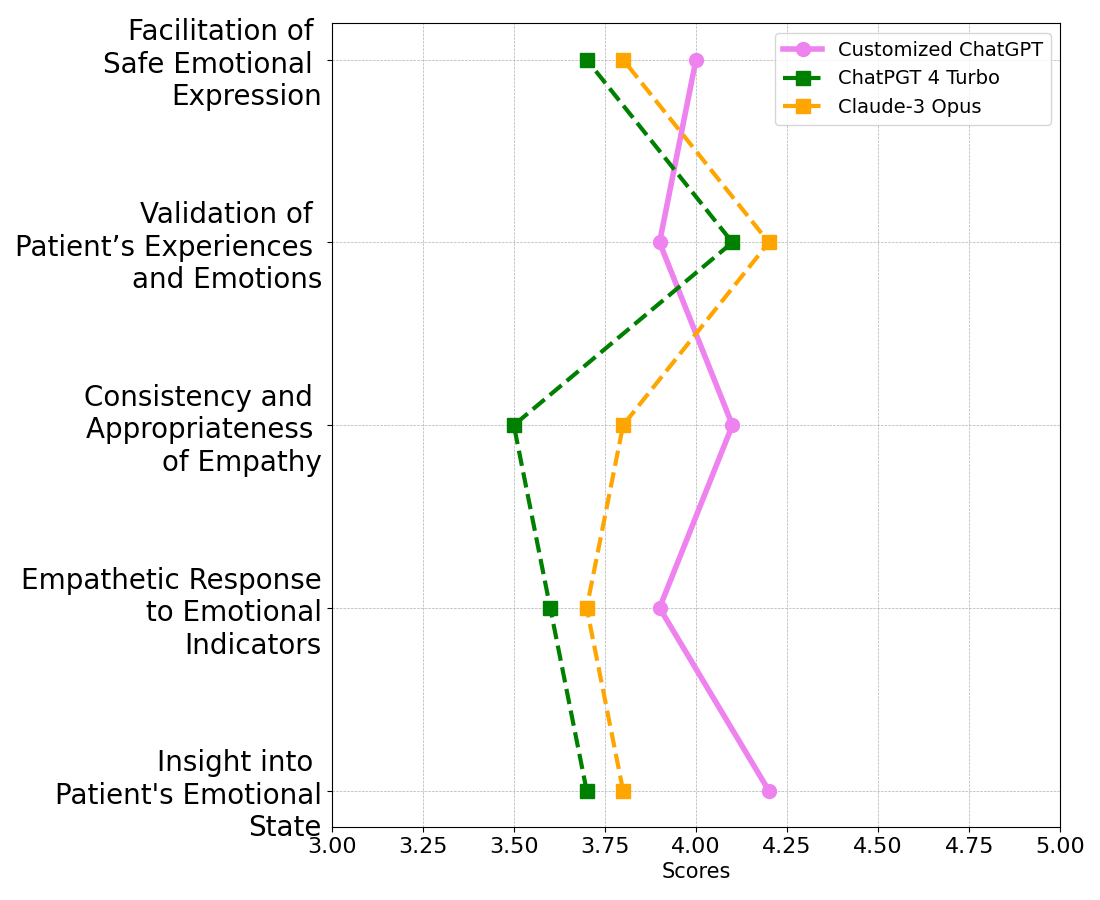}
	\caption{Comparative analysis of customized ChatGPT, ChatGPT-4 Turbo, and Claude-3 Opus on emotional understanding and empathy. This graph presents the performance across five key emotional metrics.}
	\label{fig:01emotional}
\end{figure}

Figure \ref{fig:02communication} compares the three models' performance across five key communication metrics in the communication capabilities analysis. ChatGPT-4 Turbo scores slightly lower across all metrics, suggesting potential areas for improvement in its communication strategies. Claude-3 Opus and the customized ChatGPT stand out, particularly in Clarity, consistency, and handling of misunderstandings, indicating their superior capacity to maintain engaging and contextually appropriate dialogue. Scores, ranging from 3.2 to 4.4, reveal each model's proficiency in delivering clear, coherent, and contextually relevant communication.

\begin{figure}[h!]
	\centering
	\includegraphics[width=1.0\linewidth]{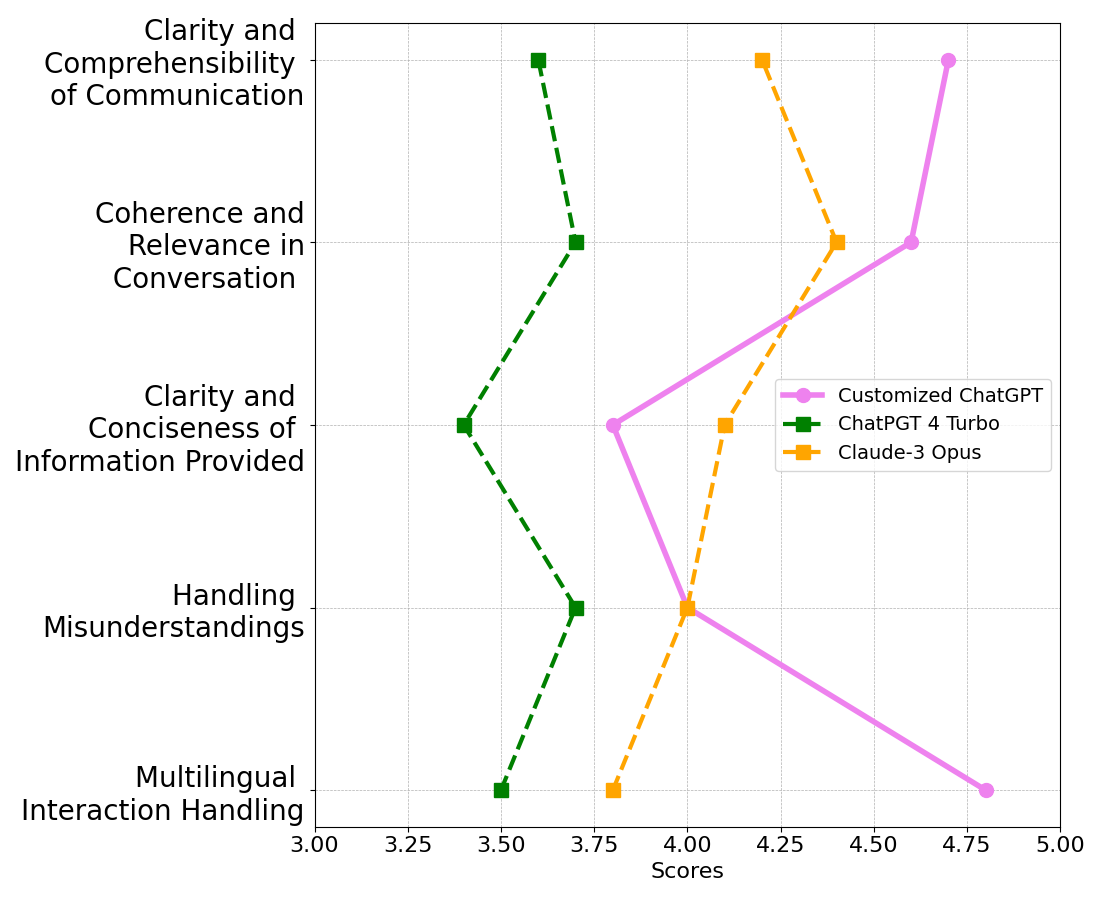}
	\caption{Comparative analysis on communication capabilities of the performance of customized ChatGPT, ChatGPT-4 Turbo, and Claude-3 Opus across five essential communication metrics.}
	\label{fig:02communication}
\end{figure}

The results depicted in Figure \ref{fig:03engagement} show the differences in how the three models engage and motivate children during simulated therapy sessions. Customized ChatGPT demonstrates robust performance, especially in encouraging autonomy and self-expression and sustaining patient interest. ChatGPT-4 Turbo shows exceptional strength in sustaining patient interest and encouraging autonomy, indicating its effectiveness in maintaining engagement over time. Claude-3 Opus stands out in the "engagement level during therapy sessions" and "promotion of active participation." This suggests that Claude-3 Opus is particularly adept at creating a positive and engaging session atmosphere, which could be attributed to its specialized tuning for empathy and user engagement. Overall, each AI model showcases distinct capabilities in enhancing the therapy experience through engagement and motivation, with Customized ChatGPT and Claude-3 Opus leading in creating a highly positive therapeutic environment.

\begin{figure}[h!]
	\centering
	\includegraphics[width=1.0\linewidth]{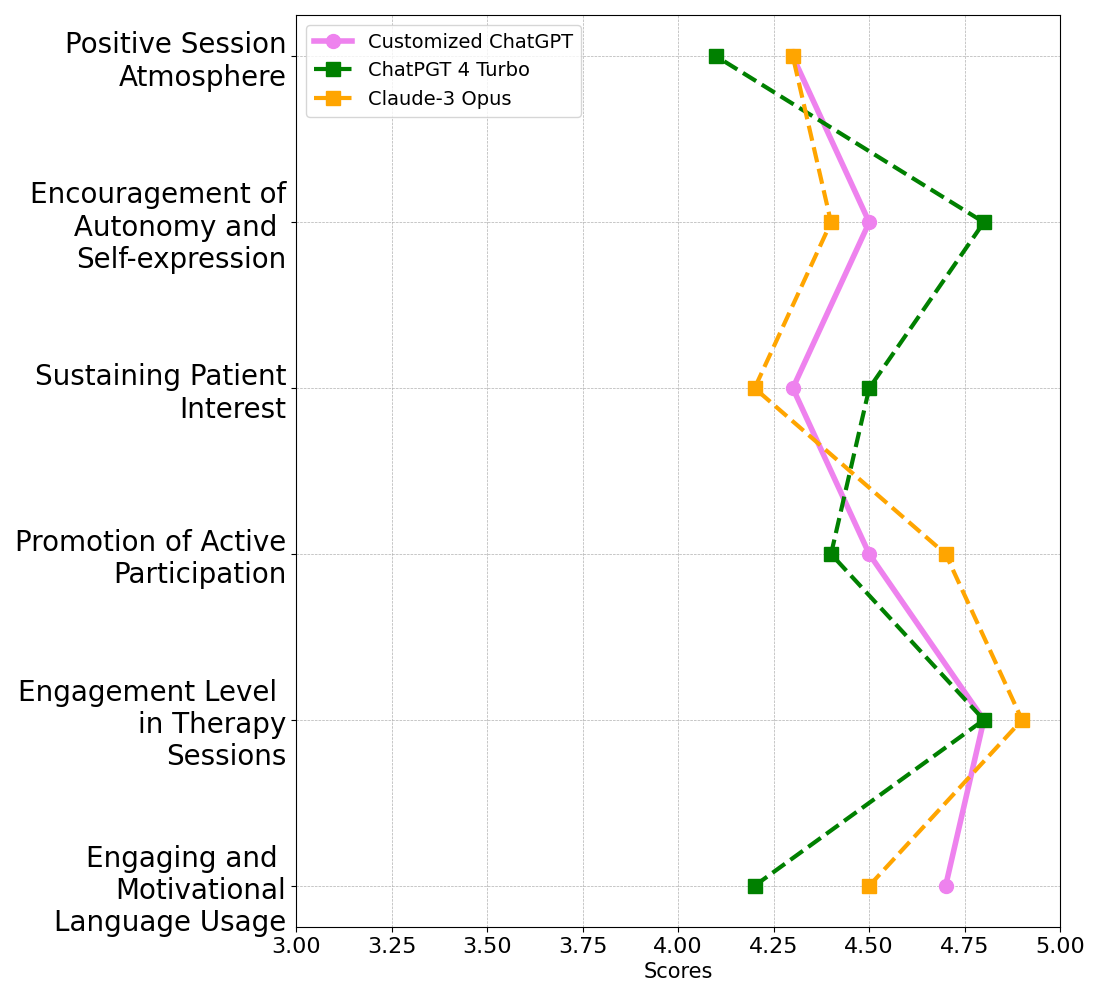}
	\caption{Comparative analysis of engagement and motivation performance of customized ChatGPT, ChatGPT-4 Turbo, and Claude-3 Opus across six metrics crucial to fostering an engaging and motivational environment in therapy sessions. }
	\label{fig:03engagement}
\end{figure}

The results depicted in Figure \ref{fig:04adaptability} present the comparative adaptability and flexibility of the three models in handling conversational dynamics. Customized ChatGPT demonstrates superior performance, particularly excelling in adjusting based on feedback, indicating its strong ability to learn and improve from interactions. This model also shows high adaptability to changing conversation dynamics. ChatGPT-4 Turbo presents more variability in its scores, with its highest in response to novel or unexpected inputs and adjustment based on feedback, suggesting it has room for improvement in adapting its conversational style and redirecting conversations. Claude-3 Opus scores consistently well across all metrics, particularly standing out in the flexibility of conversational style and adjustment based on feedback, indicating its robust capacity for handling diverse and dynamic conversational scenarios. Overall, the data highlights the nuanced capabilities of each AI model in adapting to and navigating the complexities of human conversation, with each model showcasing particular strengths in different areas of conversational adaptability and flexibility.

\begin{figure}[h!]
	\centering
	\includegraphics[width=1.0\linewidth]{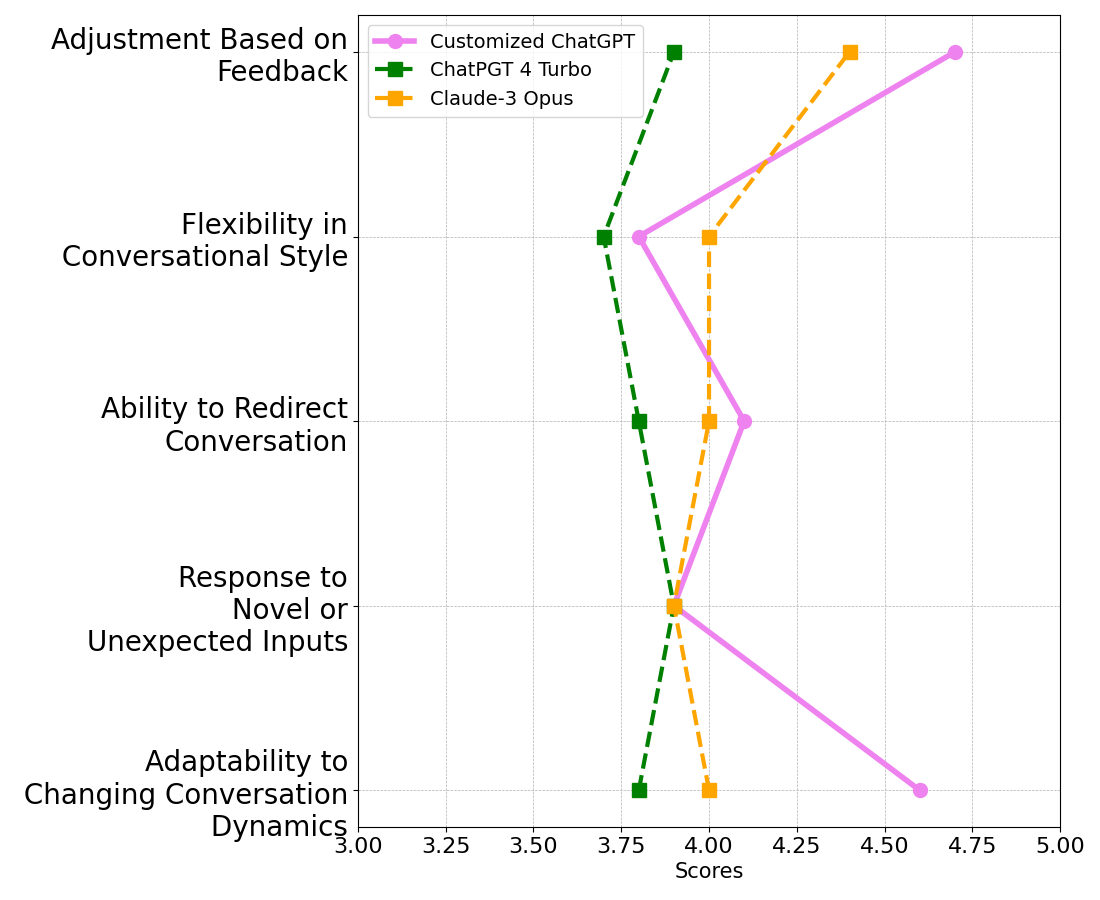}
	\caption{Comparison analysis of adaptability and flexibility performance of customized ChatGPT, ChatGPT-4 Turbo, and Claude-3 Opus across five key adaptability and flexibility metrics.}
	\label{fig:04adaptability}
\end{figure}

Figure \ref{fig:05therapy} compares the three models across nine more advanced therapeutic metrics. Notably, customized ChatGPT scores highly in areas such as "Potential for Future Applications" and "Building Trust with Patient," suggesting a solid capability for innovation and forging patient trust. On the other hand, Claude-3 Opus demonstrates superior performance in "Creation of a Safe Environment" and "Respect for Patient's Boundaries," highlighting its commitment to creating a secure and respectful therapeutic space. ChatGPT-4 Turbo displays consistent results, reflecting its balanced approach across the range of metrics. The distinct performance patterns of each model highlight their unique strengths and potential areas for development, with implications for tailoring AI tools to enhance therapeutic outcomes and patient experiences in mental health care.

\begin{figure}[h!]
	\centering
	\includegraphics[width=1.0\linewidth]{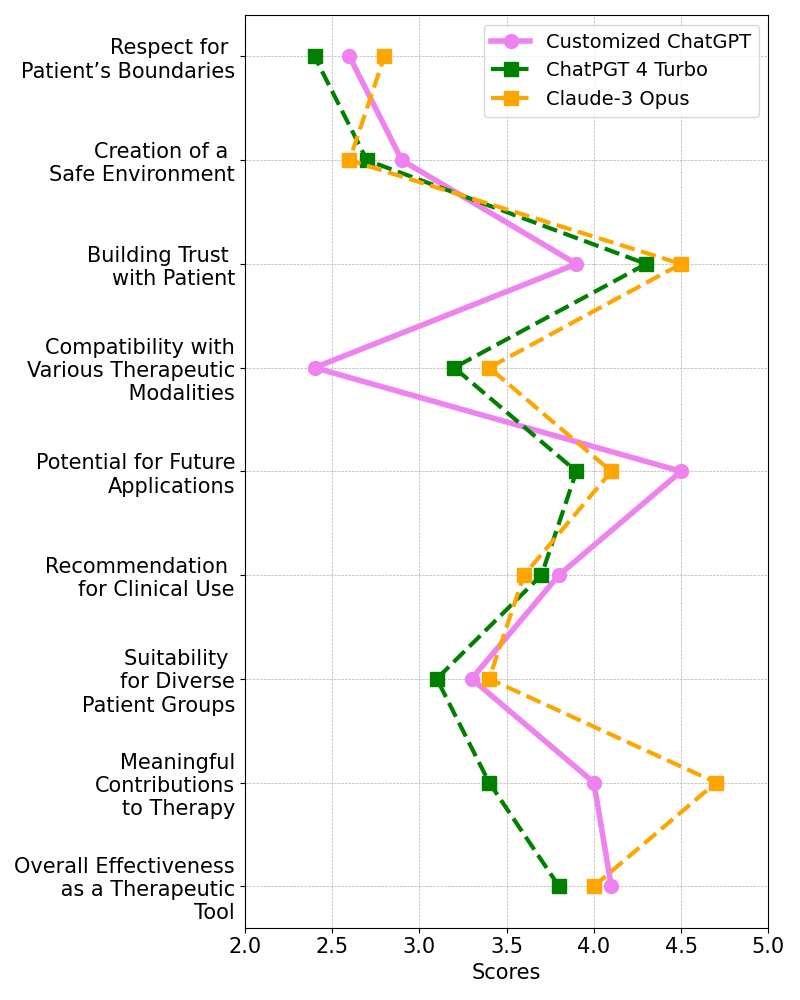}
	\caption{Comparative evaluation of the performance of customized ChatGPT, ChatGPT-4 Turbo, and Claude-3 Opus across nine therapeutic effectiveness metrics.}
	\label{fig:05therapy}
\end{figure}

\section{Discussion}\label{D}

It is essential to acknowledge that while the interactions were not real-life therapy sessions, the conclusions drawn are still within a validated framework. The simulations were rigorously reviewed and endorsed by experts in ADHD therapy. These specialists possess extensive experience and a deep understanding of the nuances and dynamics of therapeutic interactions. Their expertise lends credibility to the simulations, ensuring that the behaviors and responses modeled by the AI-powered robotic assistant closely mirror what could be expected in actual therapeutic settings. Therefore, despite the simulated environment, the findings provide valuable insights and have practical implications for the development and refinement of AI in mental health therapies.

Therefore, we emphasize the nuanced performance of the integrated ChatGPT-4 Turbo and Claude-3 Opus models within a robotic assistant framework for ADHD therapy. The results highlight the critical role of specialized training and design in developing AI with advanced emotional intelligence capabilities. The variations in performance across different models underscore the importance of tailoring AI functionalities to meet the specific demands of therapeutic sessions, balancing rapid response capabilities with the necessity for ethical and safe interactions. This balance is crucial in optimizing therapeutic engagement and ensuring practical and respectful patient care.

The integration of ChatGPT-4 Turbo and Claude-3 Opus into the robotic assistant reflects a significant advancement in ADHD therapy. ChatGPT-4 Turbo's superior performance and responsiveness suggest its potential for real-time applications, while Claude-3 Opus's understanding, coherence, and ethical focus offer a framework for safe and engaging interactions. This paper posits that the fusion of these LLMs with robotic assistants could revolutionize ADHD therapy, making sessions more engaging and effective. It calls for continued innovation and customization in AI to address the nuanced needs of ADHD therapy, proposing a future where therapeutic interventions are personalized and interactive. This discussion lays the groundwork for future challenges, particularly in fully integrating customized models like ChatGPT into robotic assistants to enhance therapy's effectiveness and engagement. The authors consider this integration a strategic enhancement to elevate smart home systems' overall functionality and user experience.

\section{Conclusions}\label{C}

This study has elucidated the transformative implications of merging advanced language models with robotic assistance in ADHD therapy. Our systematic approach, encompassing both technical and clinical evaluations, has revealed that the integration of ChatGPT-4 Turbo within the robotic assistant advances the interactivity of therapeutic sessions, mainly due to its rapid response and high performance. Claude-3 Opus has emerged as an empathetic model that guarantees safe user interactions, essential for the delicate therapeutic environment of ADHD.

By exploring these models, our findings advocate for a more personalized, engaging approach to therapy. This is underpinned by the capacity of AI to adapt to individual needs, as well as by the efficacy of a robotic medium that can consistently engage children. The evidence from this research suggests significant strides in making therapy more accessible and attuned to the unique challenges faced by individuals with ADHD.

Looking ahead, our continued development aims to embrace the full potential of AI, specifically to analyze behavioral patterns and enhance the strategic planning of ADHD therapies. While the field awaits regulatory developments to catch up with technological advancements, our dedication to progress remains steadfast. We are committed to refining these models to further support Socially Assistive Robots in therapeutic settings, ensuring that children with ADHD receive the most innovative care possible.

\bibliography{References}

\begin{thebibliography}{10}

\bibitem{fletcher2021attention}
J.~M. Fletcher, ``Attention-deficit/hyperactivity disorder (adhd),'' {\em
  Developmental Psychopathology}, pp.~89--118, 2021.

\bibitem{koutsoklenis2023adhd}
A.~Koutsoklenis and J.~Honkasilta, ``Adhd in the dsm-5-tr: What has changed and
  what has not,'' {\em Frontiers in psychiatry}, vol.~13, p.~1064141, 2023.

\bibitem{bell2011critical}
A.~S. Bell, ``A critical review of adhd diagnostic criteria: What to address in
  the dsm-v,'' {\em Journal of Attention Disorders}, vol.~15, no.~1, pp.~3--10,
  2011.

\bibitem{lambez2020non}
B.~Lambez, A.~Harwood-Gross, E.~Z. Golumbic, and Y.~Rassovsky,
  ``Non-pharmacological interventions for cognitive difficulties in adhd: A
  systematic review and meta-analysis,'' {\em Journal of psychiatric research},
  vol.~120, pp.~40--55, 2020.

\bibitem{healthcare11202795}
A.-M. Gabaldón-Pérez, M.-L. Martín-Ruiz, F.~Díez-Muñoz, M.~Dolón-Poza,
  N.~Máximo-Bocanegra, and I.~Pau de~la Cruz, ``The potential of digital
  screening tools for childhood adhd in school environments: A preliminary
  study,'' {\em Healthcare}, vol.~11, no.~20, 2023.

\bibitem{westwood2023computerized}
S.~J. Westwood, V.~Parlatini, K.~Rubia, S.~Cortese, and E.~J. Sonuga-Barke,
  ``Computerized cognitive training in attention-deficit/hyperactivity disorder
  (adhd): a meta-analysis of randomized controlled trials with blinded and
  objective outcomes,'' {\em Molecular Psychiatry}, vol.~28, no.~4,
  pp.~1402--1414, 2023.

\bibitem{oladimeji2023impact}
K.~E. Oladimeji, A.~Nyatela, S.~Gumede, D.~Dwarka, and S.~T. Lalla-Edward,
  ``Impact of artificial intelligence (ai) on psychological and mental health
  promotion: An opinion piece,'' {\em New Voices in Psychology}, vol.~13,
  pp.~12--pages, 2023.

\bibitem{darzi2023could}
P.~Darzi, ``Could artificial intelligence be a therapeutic for mental
  issues?,'' {\em Science Insights}, vol.~43, no.~5, pp.~1111--1113, 2023.

\bibitem{achiam2023gpt}
J.~Achiam, S.~Adler, S.~Agarwal, L.~Ahmad, I.~Akkaya, F.~L. Aleman, D.~Almeida,
  J.~Altenschmidt, S.~Altman, S.~Anadkat, {\em et~al.}, ``Gpt-4 technical
  report,'' {\em arXiv preprint arXiv:2303.08774}, 2023.

\bibitem{cowen2024claude}
T.~Cowen, ``Claude 3 opus and agi,'' 2024.

\bibitem{cheng2023now}
S.-W. Cheng, C.-W. Chang, W.-J. Chang, H.-W. Wang, C.-S. Liang, T.~Kishimoto,
  J.~P.-C. Chang, J.~S. Kuo, and K.-P. Su, ``The now and future of chatgpt and
  gpt in psychiatry,'' {\em Psychiatry and clinical neurosciences}, vol.~77,
  no.~11, pp.~592--596, 2023.

\bibitem{rabbitt2015integrating}
S.~M. Rabbitt, A.~E. Kazdin, and B.~Scassellati, ``Integrating socially
  assistive robotics into mental healthcare interventions: Applications and
  recommendations for expanded use,'' {\em Clinical psychology review},
  vol.~35, pp.~35--46, 2015.

\bibitem{Tamdjidi1778288}
R.~Tamdjidi and D.~Pagès~Billai, ``Chatgpt as an assistive technology to
  enhance reading comprehension for individuals with adhd,''

\bibitem{estevez2021case}
D.~Est{\'e}vez, M.-J. Terr{\'o}n-L{\'o}pez, P.~J. Velasco-Quintana, R.-M.
  Rodr{\'\i}guez-Jim{\'e}nez, and V.~{\'A}lvarez-Manzano, ``A case study of a
  robot-assisted speech therapy for children with language disorders,'' {\em
  Sustainability}, vol.~13, no.~5, p.~2771, 2021.

\bibitem{amato2021socially}
F.~Amato, M.~Di~Gregorio, C.~Monaco, M.~Sebillo, G.~Tortora, and G.~Vitiello,
  ``Socially assistive robotics combined with artificial intelligence for
  adhd,'' in {\em 2021 IEEE 18th Annual Consumer Communications \& Networking
  Conference (CCNC)}, pp.~1--6, IEEE, 2021.

\bibitem{lai2021data}
Y.~H. Lai, Y.~C. Chang, C.~W. Tsai, C.~H. Lin, and M.~Y. Chen, ``Data fusion
  analysis for attention-deficit hyperactivity disorder emotion recognition
  with thermal image and internet of things devices,'' {\em Software: Practice
  and Experience}, vol.~51, no.~3, pp.~595--606, 2021.

\bibitem{arpaia2020wearable}
P.~Arpaia, L.~Duraccio, N.~Moccaldi, and S.~Rossi, ``Wearable brain--computer
  interface instrumentation for robot-based rehabilitation by augmented
  reality,'' {\em IEEE Transactions on instrumentation and measurement},
  vol.~69, no.~9, pp.~6362--6371, 2020.

\bibitem{rakhymbayeva2020long}
N.~Rakhymbayeva, N.~Seitkazina, D.~Turabayev, A.~Pak, and A.~Sandygulova, ``A
  long-term study of robot-assisted therapy for children with severe autism and
  adhd,'' in {\em Companion of the 2020 ACM/IEEE International Conference on
  Human-Robot Interaction}, pp.~401--402, 2020.

\bibitem{zhanatkyzy2020quantitative}
A.~Zhanatkyzy, Z.~Telisheva, A.~Turarova, Z.~Zhexenova, and A.~Sandygulova,
  ``Quantitative results of robot-assisted therapy for children with autism,
  adhd and delayed speech development,'' in {\em Companion of the 2020 ACM/IEEE
  International Conference on Human-Robot Interaction}, pp.~541--542, 2020.

\bibitem{kumazaki2021enhancing}
H.~Kumazaki, T.~Muramatsu, Y.~Yoshikawa, H.~Haraguchi, T.~Sono, Y.~Matsumoto,
  H.~Ishiguro, M.~Kikuchi, T.~Sumiyoshi, and M.~Mimura, ``Enhancing
  communication skills of individuals with autism spectrum disorders while
  maintaining social distancing using two tele-operated robots,'' {\em
  Frontiers in psychiatry}, vol.~11, p.~1641, 2021.

\bibitem{berrezueta2021assessment}
J.~Berrezueta-Guzman, I.~Pau, M.-L. Mart{\'\i}n-Ruiz, and
  N.~M{\'a}ximo-Bocanegra, ``Assessment of a robotic assistant for supporting
  homework activities of children with adhd,'' {\em IEEE Access}, vol.~9,
  pp.~93450--93465, 2021.

\bibitem{vita2019neurobot}
S.~Vita and A.~Mennitto, ``Neurobot: a psycho-edutainment tool to perform
  neurofeedback training in children with adhd.,'' in {\em PSYCHOBIT}, 2019.

\bibitem{berrezueta2021robotic}
J.~Berrezueta-Guzman, V.~E. Robles-Bykbaev, I.~Pau, F.~Pes{\'a}ntez-Avil{\'e}s,
  and M.-L. Mart{\'\i}n-Ruiz, ``Robotic technologies in adhd care: Literature
  review,'' {\em IEEE Access}, vol.~10, pp.~608--625, 2021.

\bibitem{berrezueta2020smart}
J.~Berrezueta-Guzman, I.~Pau, M.-L. Mart{\'\i}n-Ruiz, and
  N.~M{\'a}ximo-Bocanegra, ``Smart-home environment to support homework
  activities for children,'' {\em IEEE Access}, vol.~8, pp.~160251--160267,
  2020.

\bibitem{berrezueta2022design}
J.~Berrezueta-Guzman, S.~Krusche, and L.~Serpa-Andrade, ``Design, development
  and assessment of a multipurpose robotic assistant in the field of cognitive
  therapy,'' {\em Human Factors in Robots, Drones and Unmanned Systems},
  vol.~57, p.~31, 2022.

\bibitem{dolon2020creation}
M.~Dol{\'o}n-Poza, J.~Berrezueta-Guzman, and M.-L. Mart{\'\i}n-Ruiz, ``Creation
  of an intelligent system to support the therapy process in children with
  adhd,'' in {\em Conference on Information and Communication Technologies of
  Ecuador}, pp.~36--50, Springer, 2020.

\bibitem{lopez2020development}
L.~L{\'o}pez-P{\'e}rez, J.~Berrezueta-Guzman, and M.-L. Mart{\'\i}n-Ruiz,
  ``Development of a home accompaniment system providing homework assistance
  for children with adhd,'' in {\em Conference on Information and Communication
  Technologies of Ecuador}, pp.~22--35, Springer, 2020.

\bibitem{berrezueta2022artificial}
J.~Berrezueta-Guzman, S.~Krusche, L.~Serpa-Andrade, and M.-L. Mart{\'\i}n-Ruiz,
  ``Artificial vision algorithm for behavior recognition in children with adhd
  in a smart home environment,'' in {\em Proceedings of SAI Intelligent Systems
  Conference}, pp.~661--671, Springer, 2022.

\bibitem{berrezueta2022user}
J.~Berrezueta-Guzman, M.-L. Martin-Ruiz, I.~Pau, and S.~Krusche, ``A
  user-centered methodology approach for the development of robotic assistants
  for pervasive unsupervised occupational therapy,'' in {\em Proceedings of the
  8th International Conference on Robotics and Artificial Intelligence},
  pp.~20--25, 2022.

\bibitem{humphrey2020delphi}
S.~Humphrey-Murto, T.~J. Wood, C.~Gonsalves, K.~Mascioli, and L.~Varpio, ``The
  delphi method,'' {\em Academic Medicine}, vol.~95, no.~1, p.~168, 2020.

\bibitem{healthcare12060683}
S.~Berrezueta-Guzman, M.~Kandil, M.-L. Martín-Ruiz, I.~Pau de~la Cruz, and
  S.~Krusche, ``Future of adhd care: Evaluating the efficacy of chatgpt in
  therapy enhancement,'' {\em Healthcare}, vol.~12, no.~6, 2024.

\bibitem{ElevenLabs2023}
{ElevenLabs}, ``Ai voice generator \& text to speech,'' 2024.
\newblock Retrieved March 22, 2024.

\end{thebibliography}
\bibliographystyle{ieeetr}

\end{document}